\documentclass[a4paper,conference]{IEEEtran}
\usepackage{cite}

\usepackage{booktabs}

\ifCLASSINFOpdf
  \usepackage[pdftex]{graphicx}
\else
\fi
\usepackage{amsmath}
\usepackage{array}

\ifCLASSOPTIONcompsoc
  \usepackage[caption=false,font=normalsize,labelfont=sf,textfont=sf]{subfig}
\else
 \usepackage[caption=false,font=footnotesize]{subfig}
\fi

\usepackage{stfloats}
\usepackage{url}

\usepackage{hyperref}
\newcommand\fnurl[2]{\href{#2}{#1}\footnote{\url{#2}}}
\newcommand{\etal} {\textit{et~al.}}
\usepackage{xcolor}

\begin{document}

\title{Unsupervised deep learning for text line segmentation}

\author{\IEEEauthorblockN{Berat Kurar Barakat\IEEEauthorrefmark{1}, 
Ahmad Droby\IEEEauthorrefmark{1}, 
Reem Alaasam\IEEEauthorrefmark{1}, 
Boraq Madi\IEEEauthorrefmark{1},\\
Irina Rabaev\IEEEauthorrefmark{4},
Raed Shammes\IEEEauthorrefmark{1} and
Jihad El-Sana\IEEEauthorrefmark{1}}

\IEEEauthorblockA{\IEEEauthorrefmark{1}
Ben-Gurion University of the Negev\\
\{berat,drobya,rym,borak,rshammes\}@post.bgu.ac.il}
\IEEEauthorblockA{\IEEEauthorrefmark{4}
Shamoon College of Engineering\\
irinar@ac.sce.ac.il}}


%


\maketitle

\begin{abstract}
We present an unsupervised deep learning method for text line segmentation that is inspired by the relative variance between text lines and spaces among text lines. Handwritten text line segmentation is important for the efficiency of further processing. A common method is to train a deep learning network for embedding the document image into an image of blob lines that are tracing the text lines. Previous methods learned such embedding in a supervised manner, requiring the annotation of many document images. This paper presents an unsupervised embedding of document image patches without a need for annotations. The number of foreground pixels over the text lines is relatively different from the number of foreground pixels over the spaces among text lines. Generating similar and different pairs relying on this principle definitely leads to outliers. However, as the results show, the outliers do not harm the convergence and the network learns to discriminate the text lines from the spaces between text lines. Remarkably, with a challenging Arabic handwritten text line segmentation dataset, VML-AHTE, we achieved superior performance over the supervised methods. Additionally, the proposed method was evaluated on the ICDAR 2017 and ICFHR 2010 handwritten text line segmentation datasets.

\end{abstract}


%
\IEEEpeerreviewmaketitle

\section{introduction}
\label{introduction}
\begin{figure*}[b]
\centering
\includegraphics[width=13cm]{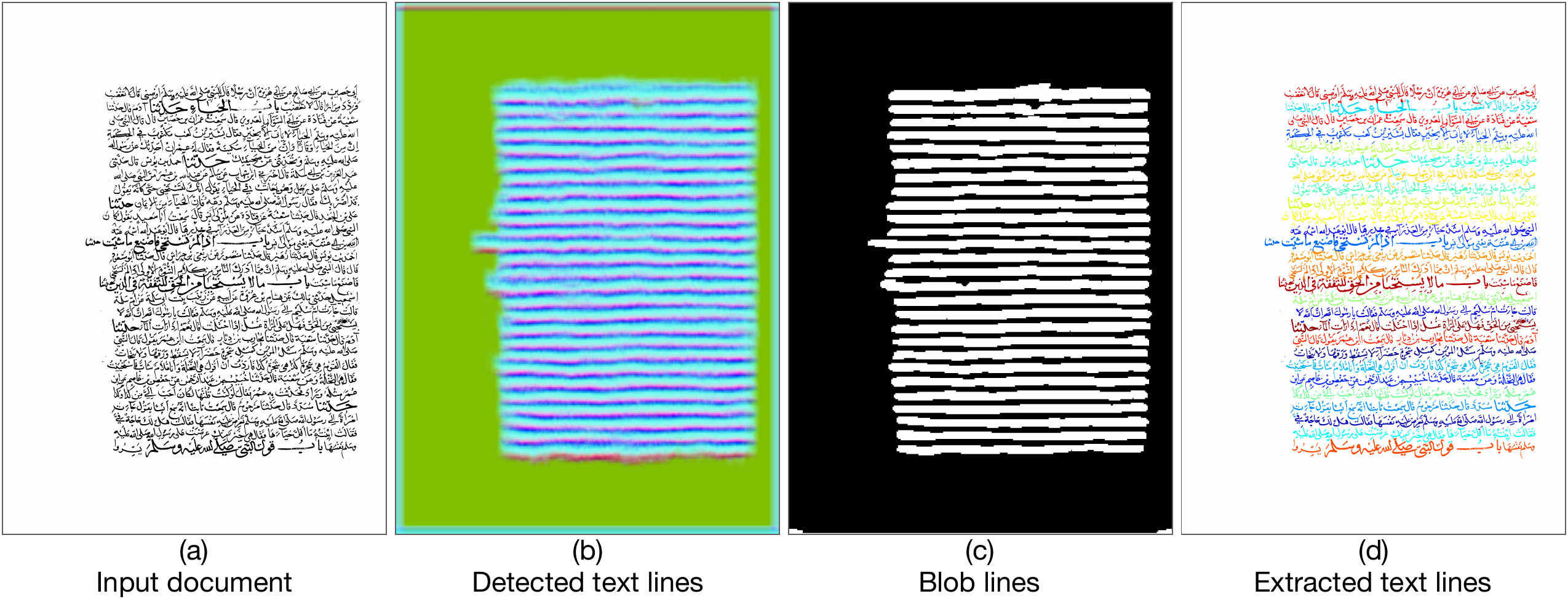}
\caption{Given a handwritten document image (a), UTLS learns to extract representation vectors of image patches where the distances between these vectors are proportional to the similarity of patches. Three principal components of patch representation vectors are visualized as a pseudo-RGB image (b). The pseudo-RGB images are thresholded onto blob lines that strike through text lines (c). Energy minimization with the assistance of detected blob lines extracts the pixel labels of text lines (d).}
\label{phases}
\end{figure*}

Text line segmentation is a classical document image analysis problem that has impact on the performance of subsequent analysis operations. The objective of text line segmentation is to recognize all the pixels that belong to a text line, as shown in \figurename~\ref{phases}(d). Text line segmentation contains both, text line detection and text line extraction. Text line detection roughly locates text line patterns, whereas text line extraction precisely assigns pixels to the text lines. Detection results can be represented by baselines or blob lines (\figurename~\ref{phases}(c)). Extraction can be represented by pixel labels (\figurename~\ref{phases}(d)) or bounding polygons. The final goal of a text line segmentation procedure is to provide text lines one by one into the next document analysis procedure.

Recently, numerous deep learning based methods have been proposed for text line segmentation of handwritten documents. Learning based methods \cite{renton2018fully,gruning2019two,oliveira2018dhsegment,barakat2018text} can inherently handle the problems arising from complex layout of text lines and heterogeneity of documents. However, they require a vast amount of labeling effort which consumes time not less than carefully designed ad-hoc heuristics \cite{li2008script,shi2009steerable,bukhari2009script,cohen2013robust}. Intuitively, labeling effort is favorable over designing ad-hoc heuristics because the former can be accomplished by human recognition skills, whereas the latter requires further mathematical skills.

This paper presents a simple but interestingly successful unsupervised convolutional network for text line segmentation. The input for the network is an unlabeled document image, and the output is segmentation of text lines. The main idea can be formulated that the visual discrimination of number of foreground pixels in document image patches requires machine to learn features that represent proximity and similarity of the elements in the document image. According to the Gestalt principle \cite{koffka2013principles}, such relevance among the elements of a document image forms the basis of unsupervised segmentation of text lines. In the first phase, we train a siamese network to learn that two document image patches with relatively same/distinct number of foreground pixels are similar/different. Certainly, this measurement assigns many pairs improperly. However, the outliers do not harm the convergence of the machine learning \cite{danon2019unsupervised}. Next, we extract representation vectors of document image patches using the penultimate layer of a single branch of the siamese network. Then, we reduce dimensions of these vectors into their three principle components, which enables producing pseudo-RGB images where similar pixels in the embedded space correspond to similar colors \cite{danon2019unsupervised}. The pseudo-RGB images are thresholded into blob lines that hover the text lines. In the last phase, text lines are labeled in pixel level using an energy minimization framework with the assistance of the detected blob lines \cite{kurar2020text}. Experiments on an Arabic handwritten textline extraction dataset, which possesses challenges by crowded and cramped text lines, show that Unsupervised Text Line Segmentation (UTLS) is more effective than supervised methods. In addition, we achieved comparable results on ICDAR 2017 \cite{simistira2017icdar2017} and ICFHR 2010 \cite{gatos2010icfhr} handwritten text line segmentation datasets.

\section{Related Work}
\label{RelatedWork}
Text line detection and segmentation in historical document images have been widely studied during the last decades, but still remains an open problem for challenging documents.

During the years, numerous methods for text line extraction  have been proposed. Among the early approaches are projection profiles based methods, which were first applied to documents with horizontal text lines~\cite{ha1995document,manmatha1999scale}, and subsequently adapted to document with skewed~\cite{arivazhagan2007statistical,bar2009line} and multi-skewed text lines~\cite{ouwayed2012general}.
Another wide class of methods are grouping or clustering methods that aggregate elements (such as pixels or connected components) in a bottom up strategy ~\cite{rabaev2013text,bukhari2009script,cohen2014using,gruuening2017robust}. Smearing based methods
~\cite{wong1982document,alaei2011new,li2008script,bukhari2009script,shi2009steerable,barakat2019vml} target to enhance the text line structure. Seam-carving methods build  energy map and compute seams that separate text lines (or seams that pierce through text lines) ~\cite{saabni2011language,asi2011text,alberti2019labeling,scius2019layout}. 
Recently, learning-based methods have shown promising results when applied for text line segmentation of handwritten documents.
Renton~\etal~\cite{renton2018fully} employed a variant of Fully Convolutional Network (FCN) with dilated convolutions for text line extraction. The model is trained to output an $X$-height pixel labeling as text line representation.
Oliveira~\etal~\cite{oliveira2018dhsegment} presented a CNN-based pixel-wise predictor for addressing multiple tasks simultaneously: page extraction, layout analysis, baseline extraction, and illustration and photograph extraction.
Their network is trained to predict the binary mask of polygonal lines that represent baselines.
Kurar~\etal~\cite{barakat2018text} build a FCN to predict text line masks. Their method targeted challenging documents, which contain curved, multi-skewed and multi-directed text lines of different fonts types and sizes.
Kiessling~\etal~\cite{kiessling2019badam} presented method based on a fully convolutional encoder-decoder network to detect baselines in document images. The baseline definition was modified slightly towards manuscripts written in Arabic scripts.
Mechi~~\etal~\cite{mechi2019text} and Neche~\etal~\cite{neche2019arabic} used an U-net and RU-net deep-learning models, which are variants of FCN. The models are trained for $X$-height based pixel-wise classifications of text lines.

All of the learning based methods reviewed above are supervised methods. We are not aware of any unsupervised deep learning approach for text line segmentation. In this paper we present an unsupervised deep learning method for text line segmentation, and evaluate it on three publicly available datasets.

\section{Method}
\label{method}
We present a method for unsupervised text line segmentation (UTLS) and show its effectiveness on handwritten document images. The method uses a siamese convolutional network to predict whether two given document image patches are similar or different, driven by the number of foreground pixels in the patches. After the training phase, a single branch of the trained network is used to extract features of document image patches, which are in turn visualized as pseudo-RGB images (\figurename~\ref{phases}(b)) and thresholded into blob lines that strike through text lines (\figurename~\ref{phases}(c)). Finally, we use an energy minimization framework \cite{barakat2019vml} to extract the pixel labels of text lines with the assistance of the detected blob lines (\figurename~\ref{phases}(d)). This section provides the details of data preparation, training, visualization of blob lines and energy minimization procedures.

\subsection{Data preparation}
\label{datapreparation}
Data preparation consists of generating patches of the size $h_p\times w_p$ pixels, cropped randomly from document images and labeling every pair of patches either similar or different. The patch height $h_p$ is estimated as three times of the average character height in the document images. The patch width $w_p$ is estimated experimentally per dataset. The labeling is done automatically using a similarity score between two patches.

Given randomly cropped two image patches, let $a_i$ be the number of foreground pixels in patch $i$ where $i\in \{1,2\}$. We define the similarity score $s$ as:
\begin{equation}
s=\frac{\min(a_1,a_2)}{\max(a_1,a_2)}
\end{equation}
Assume that $a_2 > a_1$ then, 
$a_1$ and $a_2$ are most similar when 
\begin{equation}
(a_2-a_1) \to 0 \text{ and } \frac{a_1}{a_2} \to 1 \text{ and in turn } s \to 1.
\end{equation}
$a_1$ and $a_2$ are most different when 
\begin{equation}
(a_2-a_1) \to \infty \text{ and } \frac{a_1}{a_2} \to 0 \text{ and in turn } s \to 0.
\end{equation}

\subsubsection{Patches similar by number of foreground pixels}
\begin{figure}[h]
\centering
\includegraphics[width=5cm]{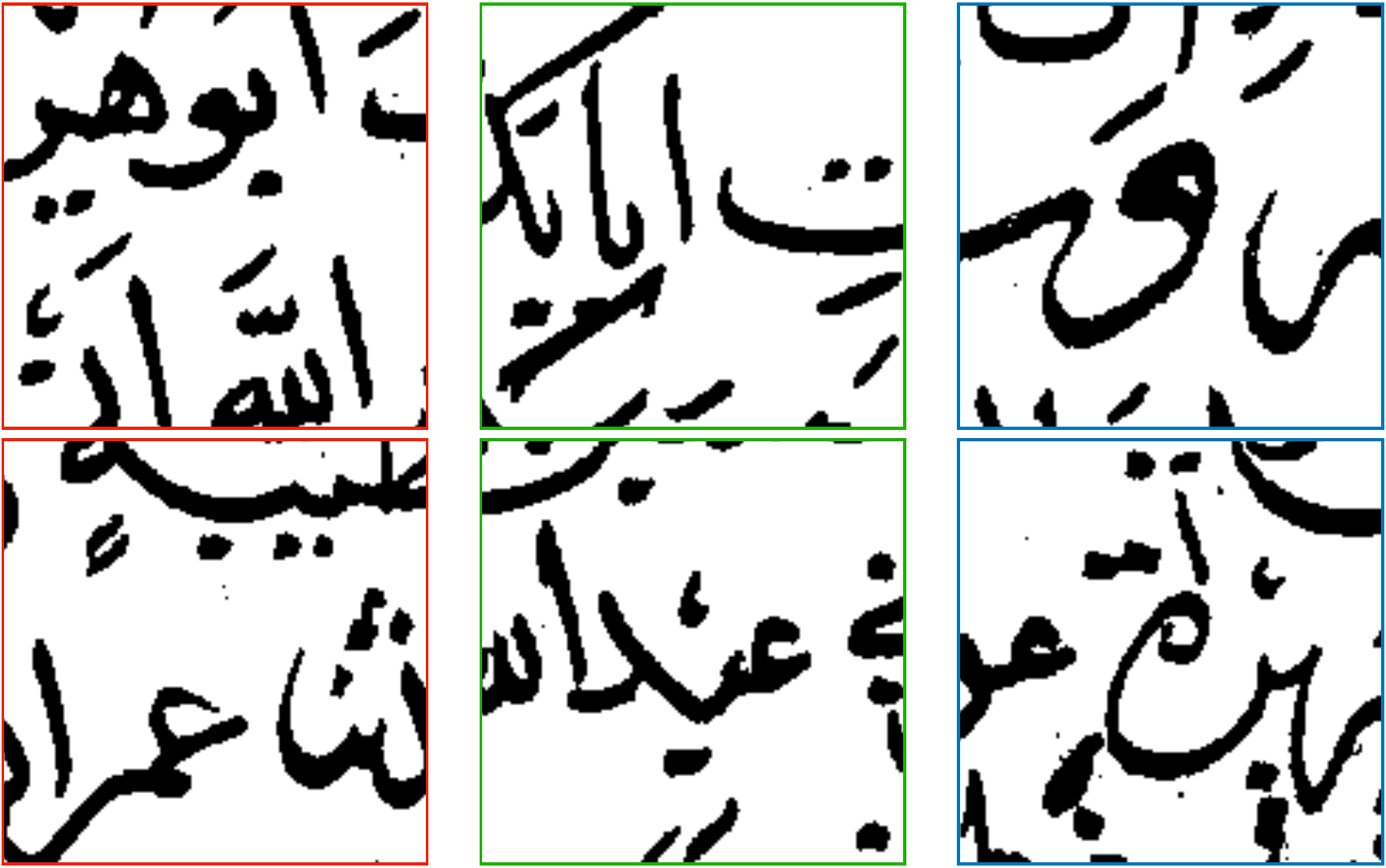}
\caption{Every column shows a pair of similar patches. In a loosely manner, both patches in each pair centralize either a text line part or a part of space between text lines.}
\label{same_pairs_by_a}
\end{figure}

This strategy continues cropping two random patches until the similarity score $s$ satisfies the following condition:
\begin{equation}
\label{similar_patches}
s\ge 0.7
\end{equation}
Intuitively this strategy generates pairs where both centralize either a text line part or a part of space between text lines (\figurename~\ref{same_pairs_by_a}). 

\subsubsection{Patches different by number of foreground pixels}
\begin{figure}[h]
\centering
\includegraphics[width=5cm]{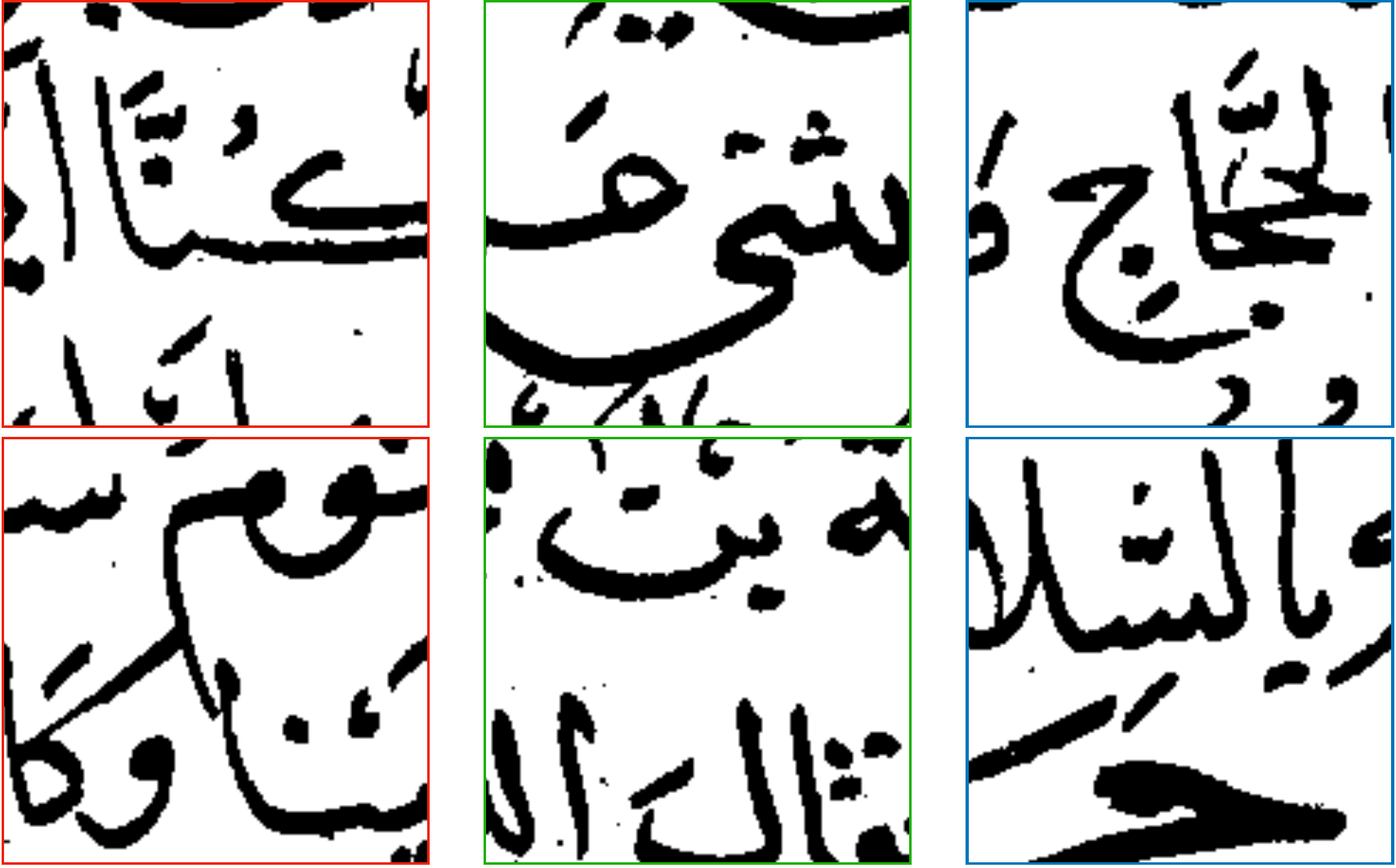}
\caption{Every column shows a pair of different patches. In a loosely manner, one of the patches in each pair centralizes a text line part and the other centralizes a part of space between text lines.}
\label{different_pairs_by_a}
\end{figure}

This strategy continues cropping two random patches until the similarity score $s$ satisfies the following condition:
\begin{equation}
\label{different_patches}
s\le 0.4
\end{equation}
Intuitively this strategy generates pairs where one centralizes a text line part and the other centralizes a part of space between text lines (\figurename~\ref{different_pairs_by_a}). 


\subsubsection{Patches different by background area}
\begin{figure}[h]
\centering
\includegraphics[width=5cm]{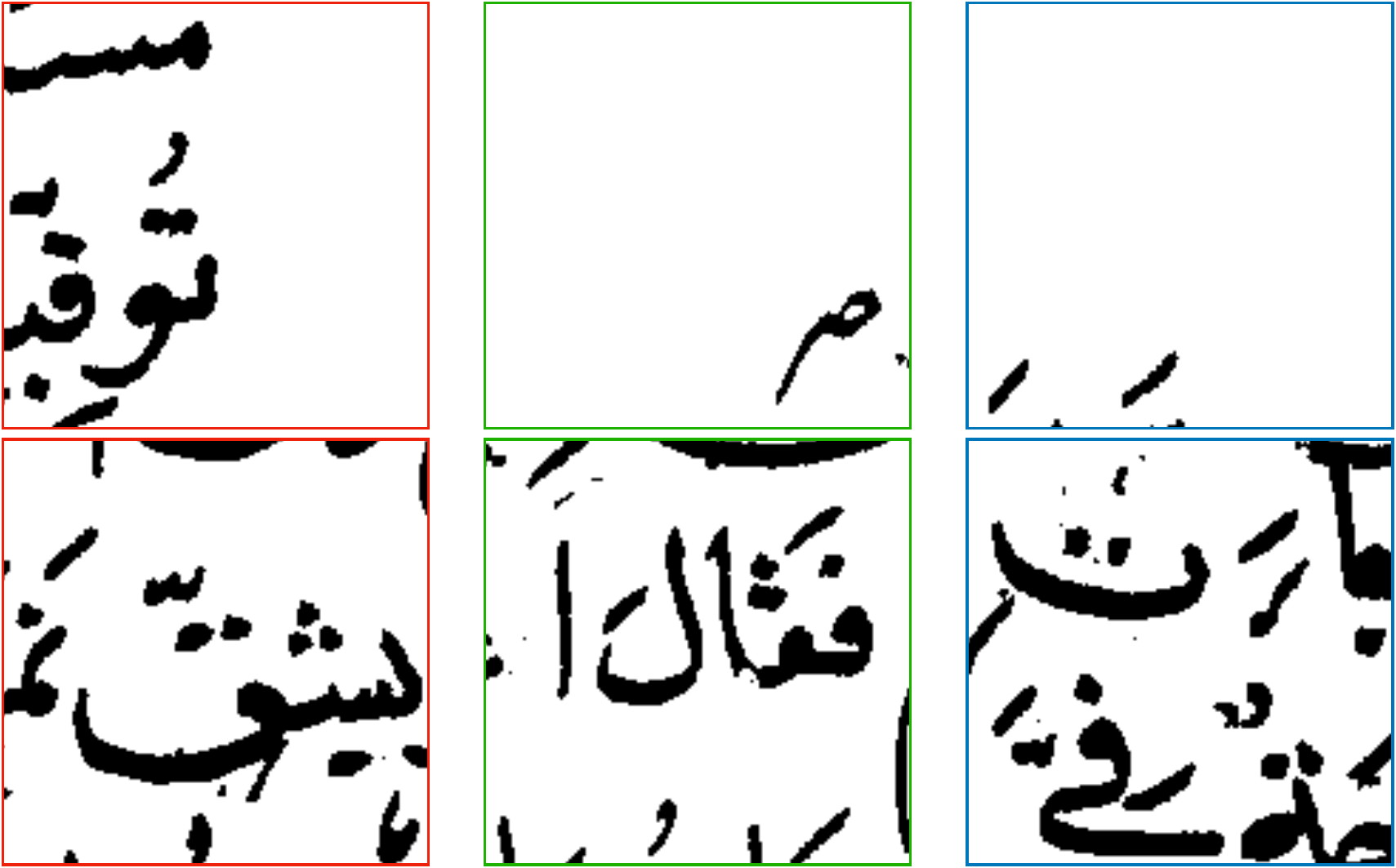}
\caption{Every column shows a pair of different patches. In a loosely manner, either of patches in each pair contains either background area or foreground area.}
\label{different_pairs_by_b}
\end{figure}

There also exist a significant difference between the background areas and the text areas in the document image. This strategy continues cropping two random patches until one of the patches is from background area and the other is from text area (\figurename~\ref{different_pairs_by_b}). We assume a patch is from background area if most of its pixels are background pixels.

\subsection{Training}
The common deep learning practice for handwritten text line segmentation is to adapt an embedding from the text lines image into a blob lines image. The classifier is first trained on a labeled set of text lines, and then expected to predict blob lines. Unlike these methods, UTLS does not need labeled data for mapping the text line image into a blob line image. It is simply trained to distinct the text lines from the spaces between text lines.

\begin{figure}[h]
\centering
\includegraphics[width=6cm]{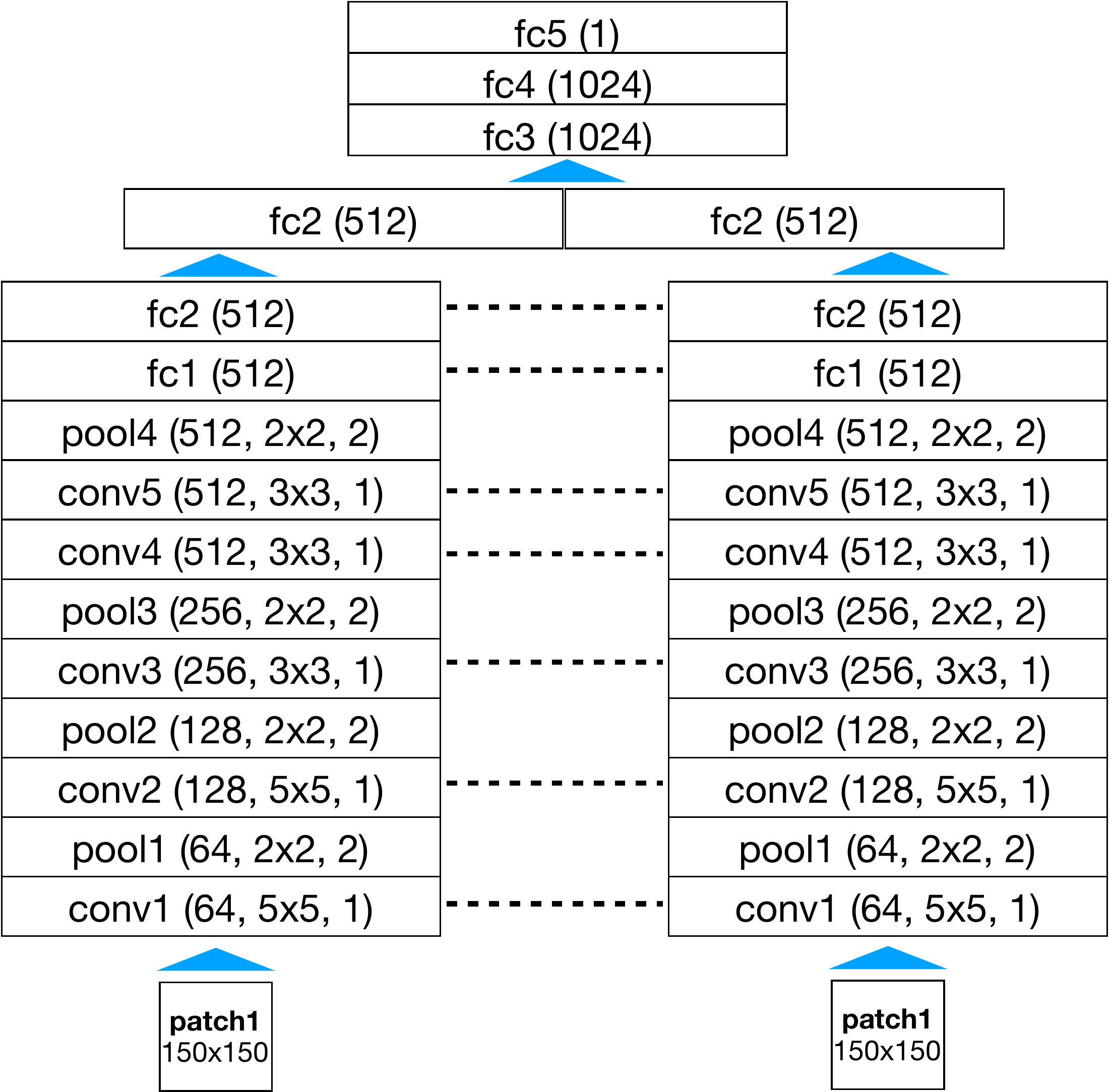}
\caption{Siamese architecture for pair similarity. Dotted lines stand for identical weights, conv stands for convolutional layer, fc stands for fully connected layer and pool is a max pooling layer.}
\label{siamese}
\end{figure}

The overall architecture is a siamese network with two identical branches. Each branch inputs an image patch and outputs a feature representation of that image patch. Consequently, these feature representations are concatenated and fed to fully connected layers in order to classify whether the two image patches are similar or different. The branches of siamese network model is based on AlexNet\cite{krizhevsky2012imagenet} and through experiments we tune the hyperparameters to fit our task. The final architecture contains two branches of CNN, each of the branches has five convolutional layers as presented in \figurename~\ref{siamese}. Dotted lines indicate identical weights, and the numbers in parentheses are the number of filters, filter size and stride. All convolutional and fully connected layers are followed by ReLU activation functions, except fc5, which feeds into a sigmoid binary classifier. The learning rate is $0.00001$ and the optimizing algorithm is ADAM.

\begin{figure}[h]
\centering
\includegraphics[width=7cm]{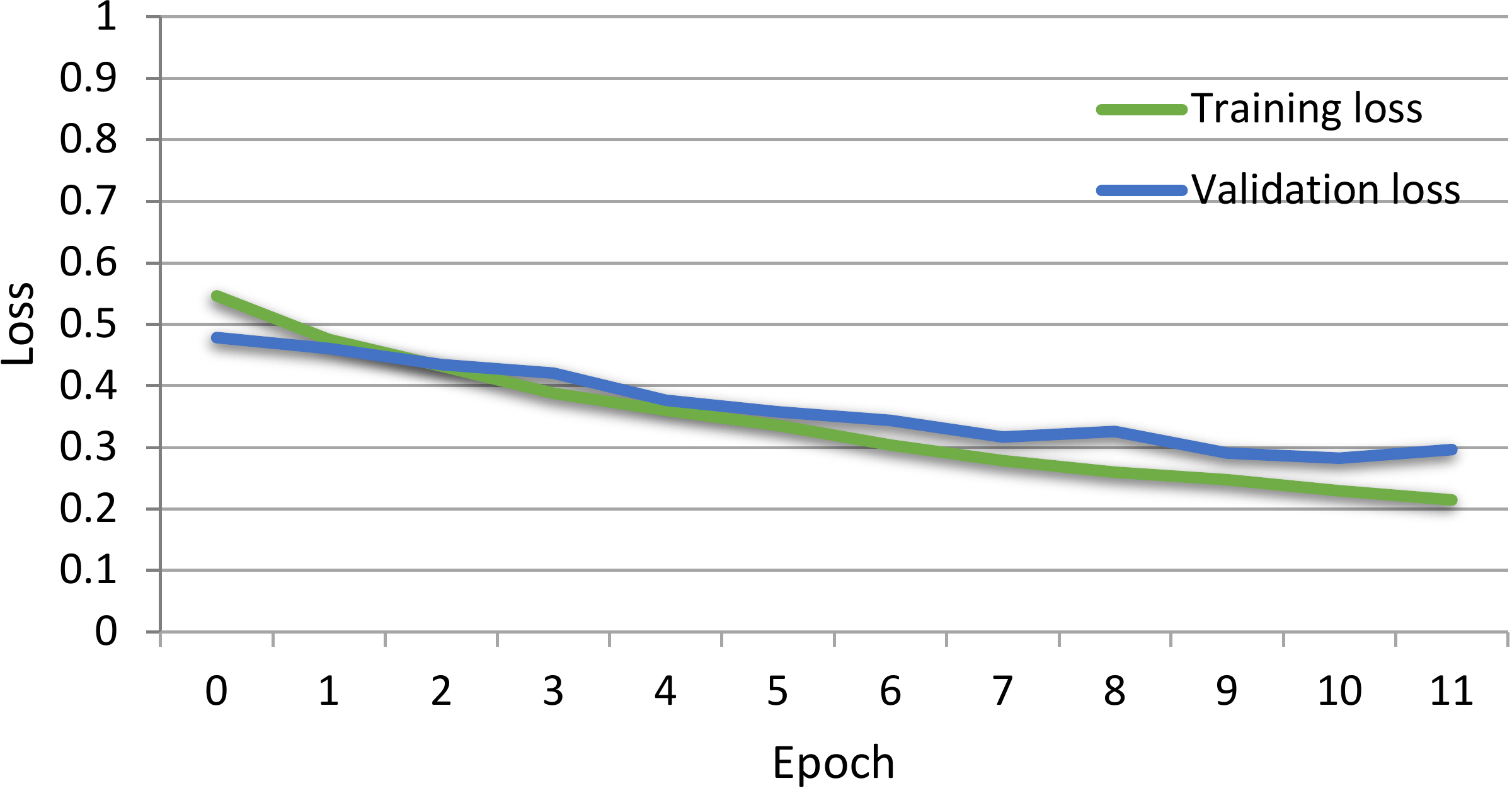}
\caption{Loss over the epochs of model training.}
\label{loss}
\end{figure}

We trained this model from scratch using $30,000$ pairs that are generated and labeled according to the strategies described in \sectionautorefname~\ref{datapreparation}, and reached a validation loss value of $0.29$ after 11 epochs (\figurename~\ref{loss}).

\subsection{Visualization of blob lines for text line detection}
Once the siamese network is trained, we use a single branch to extract the features of patches. This embeds every patch into a feature vector of $512$ dimensions. To visualize the features of a complete document image, a sliding window of the size $h_p\times w_p$ is used, but only the inner window of the size $h_i\times w_i$ is considered to eliminate the edge affect. In our experiments we find $(h_i\times w_i)=(10,10)$ gives the best results on all the datasets. We also pad the document image with white pixels at its right and bottom sides if its size is not an integer multiple of the sliding window size. An additional padding is added at 4 sides of the document image for considering only the central part of the sliding window. As a result, a document image with the size $h_d\times w_d$ is mapped to a representation matrix of the size $h_d\times w_d\times 512$. We project $512D$ vectors into their three principle components and use these components to construct pseudo-RGB image in which similar patches are assigned the similar colors (\figurename~\ref{phases}(b)). Binary blob lines image is an outcome of thresholded pseudo-RGB image (\figurename~\ref{phases}(c)).

\subsection{Energy minimization for text line extraction}
We adopt the energy minimization framework \cite{boykov2001fast} that uses graph cuts to approximate the minima of an arbitrary function. We adapt the energy function to be used with connected components for extracting the text lines. Minimum of the adapted function correspond to a good extraction which urges to assign components to the label of the closest blob line while straining to assign closer components to the same label.

Let $\mathcal{L}$ be the set of binary blob lines, and $\mathcal{C}$ be the set of components in the binary document image. Energy minimization finds a labeling $f$ that assigns each component $c\in \mathcal{C}$ to a label $l_c\in \mathcal{L}$, where energy function $\textbf{E}(f)$ has the minimum.
\begin{equation}
    \textbf{E}(f) = \sum_{c\in {\mathcal C}}D(c, \ell_c)+\sum_{\{c,c'\}\in \mathcal N}d(c, c')\cdot \delta (\ell_c \neq \ell_{c'})
\label{eq:em}
\end{equation}

The term $D$ is the data cost, $d$ is the smoothness cost, and  $\delta$ is an indicator function. Data cost is the cost of assigning component $c$ to label $l_c$. 
$D(c, \ell_c)$ is defined to be the Euclidean distance between the centroid of the component $c$ and the nearest neighbour pixel in blob line $l_c$ for the centroid of the component $c$. 
Smoothness cost is the cost of assigning neighbouring elements to different labels. Let $\mathcal{N}$ be the set of nearest component pairs. Then $\forall \{c,c'\}\in \mathcal {N}$
\begin{equation}
    d(c,c') = \exp({-\beta\cdot d_c(c,c')})
\label{eq:sc}
\end{equation}
where $d_c(c,c')$ is the Euclidean distance between the centroids of the components $c$ and $c'$,  and $\beta$ is defined as
\begin{equation}
     \beta=(2\left<d_c(c,c')\right>)^{-1}
\end{equation}
$\left<\cdot\right>$ denotes expectation over all pairs of neighbouring components \cite{boykov2001interactive} in a document page image. $\delta (\ell_c \neq \ell_{c'})$ is equal to $1$ if the condition inside the parentheses holds and $0$ otherwise.

\section{Datasets}
\label{datasets}

We evaluated the proposed method on three publicly available handwritten datasets: VML-AHTE, ICDAR 2017 \cite{simistira2017icdar2017} and ICFHR 2010 \cite{gatos2010icfhr}.

\subsection{VML-AHTE}
\begin{figure}[h]
\centering
\includegraphics[width=7cm]{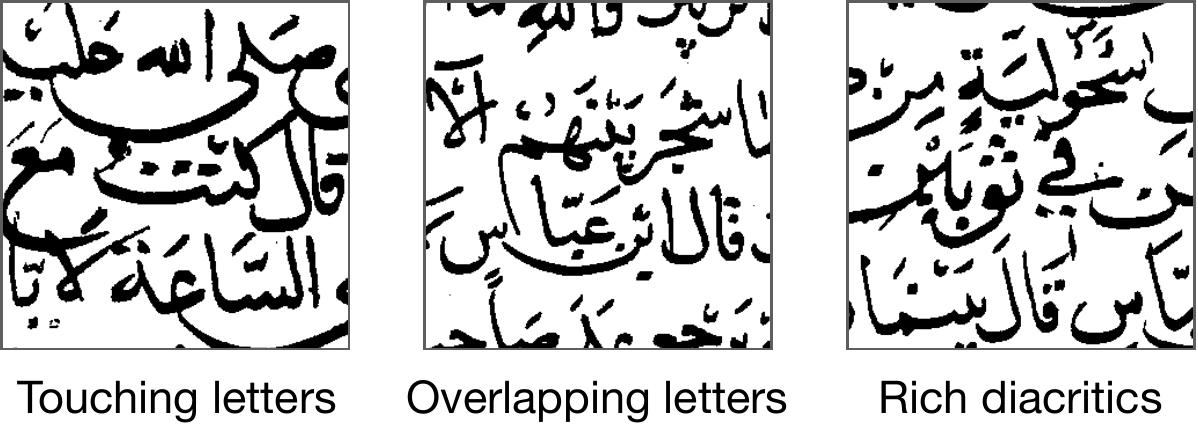}
\caption{Some samples of challenges in VML-AHTE dataset.}
\label{ahte_dataset}		
\end{figure}

Visual Media Lab - Arabic Handwritten Textline Extraction (VML-AHTE) dataset is a collection of 30 pages selected from several manuscripts. It is a newly published dataset and available online for \fnurl{downloading}{https://www.cs.bgu.ac.il/~berat/data/ahte_dataset}. VML-AHTE dataset is challenging in terms of rich diacritics, and touching and overlapping characters, as shown in Fig.\ref{ahte_dataset}.

\subsection{ICDAR 2017}
\begin{figure}[h]
\centering
\includegraphics[width=7cm]{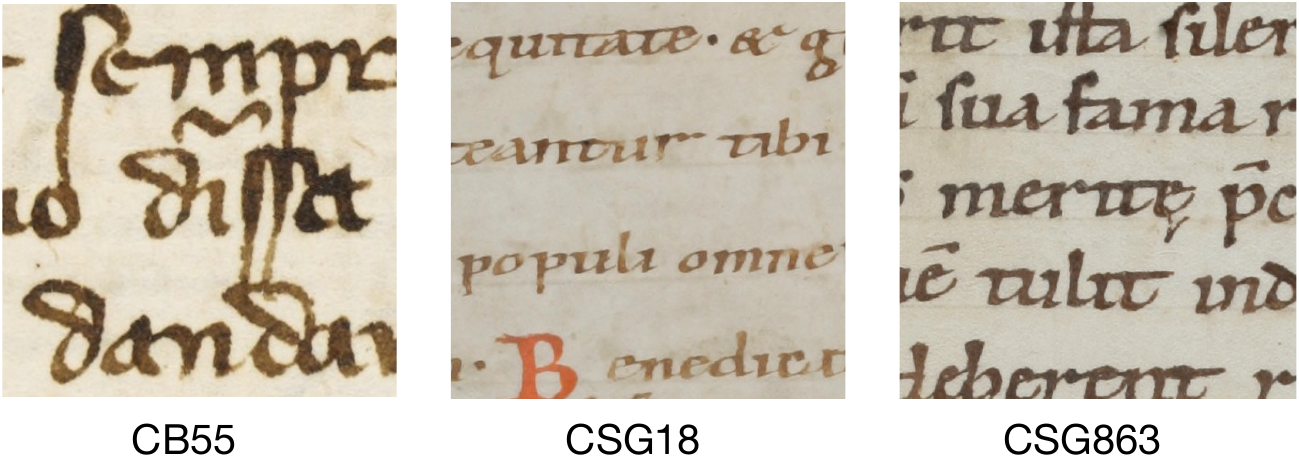}
\caption{Diva-HisDB dataset contains 3 manuscripts: CB55, CSG18 and CSG863. Notice the touching characters among multiple consecutive text lines in CB55.} 
\label{diva_dataset}
\end{figure}

ICDAR 2017 dataset \cite{simistira2017icdar2017} contains 150 pages from 3 medieval manuscripts: CB55, CSG18 and CSG863, see Fig.~\ref{diva_dataset} for an example. Among them, CB55 is characterized by a vast number of touching text lines.

\subsection{ICFHR 2010}
\begin{figure}[h]
\centering
\includegraphics[width=7cm]{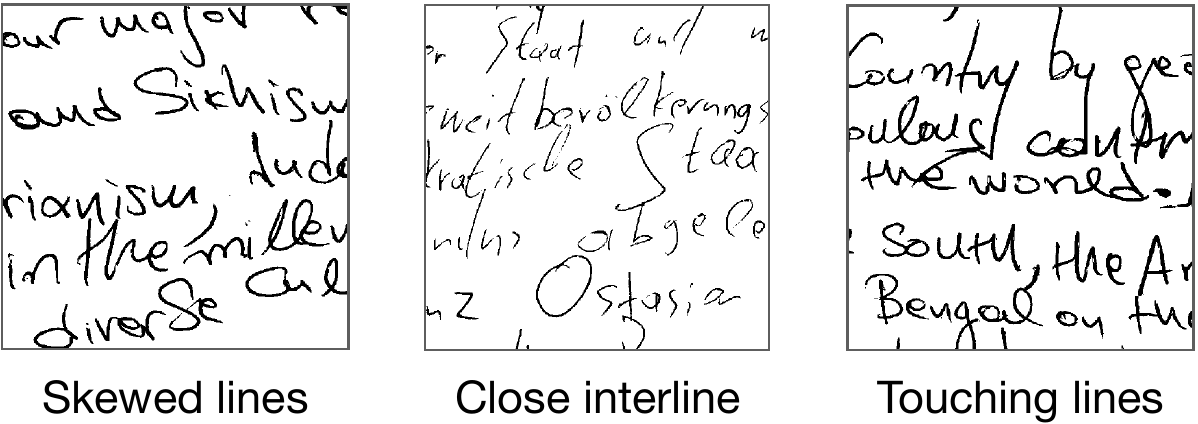}
\caption{ICFHR 2010 dataset contains unconstrained handwritten documents.}
\label{icfhr_dataset}		
\end{figure}

ICFHR 2010 dataset \cite{gatos2010icfhr} is particularly challenging as it comprises handwriting from different languages and writers. The text lines are skewed and have varying sizes as well as interline spacing, as shown in the example page in Fig.~\ref{icfhr_dataset}. 

\section{Experiments}
\label{experiments}

Our experimental study covers three datasets that are different in terms of the text line segmentation challenges they contain. On one hand, VML-AHTE dataset exhibits crowded diacritics and cramped text lines, whereas ICDAR 2017 dataset contains consequently touching text lines. On the other hand, ICFHR 2010 dataset is heterogeneous by document resolutions, text line heights and skews. Therefore, the proposed algorithm does not use universal values for all the experimented datasets. In this section we present the effect of patch size and similarity score values on the method's performance. The performance is measured using the line segmentation evaluation metrics of ICDAR 2013 \cite{gatos2010icfhr} and ICDAR 2017 \cite{alberti2017open}.

\subsection{ICDAR 2013 line segmentation evaluation metrics}
ICDAR 2013 metrics calculate recognition accuracy ($RA$), detection rate ($DR$) and F-measure ($FM$) values. Given a set of image points $I$, let $R_i$ be the set of points inside the $i^{th}$ result region, $G_j$ be the set of points inside the $j^{th}$ ground truth region, and $T(p)$ is a function that counts the points inside the set $p$, then the $MatchScore(i,j)$ is calculated by \equationautorefname~\ref{match}
\begin{equation}
  MatchScore(i,j) = \frac{T(G{j}\cap R{i})}{T(G{j}\cup R{i})} 
 \label{match}
\end{equation}
The evaluator considers a region pair $(i,j)$ as a one-to-one match if the  $MatchScore(i,j)$ is equal or above the threshold, which we set to $90$ for all evaluations except for ICFHR 2010 dataset to $95$ for results to be comparable. Let $N_1$ and $N_2$ be the number of ground truth and output elements, respectively, and let $M$ be the number of one-to-one matches. The evaluator calculates the $DR$, $RA$ and $FM$ as follows:
\begin{equation}
  DR = \frac{M}{N_1} 
\end{equation}
\begin{equation}
  RA = \frac{M}{N_2} 
\end{equation} 
\begin{equation}
  FM=\frac{2\times DR\times RA}{DR+RA} 
\end{equation}
 
\subsection{ICDAR 2017 line segmentation evaluation metrics}
ICDAR 2017 metrics are based on the Intersection over Union (IU). IU scores for each possible pair of Ground Truth (GT) polygons and Prediction (P) polygons are computed as follows: 
\begin{equation}\label{iu}
    IU=\frac{IP}{UP}
\end{equation}
IP denotes the number of intersecting foreground pixels among the pair of polygons. UP denotes number of foreground pixels in the union of foreground pixels of the pair of polygons. The pairs with maximum IU score are selected as the matching pairs of GT polygons and P polygons. Then, pixel IU and line IU are calculated among these matching pairs. For each matching pair, line TP, line FP and line FN are given by:
\begin{itemize}
    \item Line TP is the number of foreground pixels that are correctly predicted in the matching pair.
    \item Line FP is the number of foreground pixels that are falsely predicted in the matching pair.
    \item Line FN is the number of false negative foreground pixels in the matching pair.
\end{itemize}

Accordingly pixel IU is: 
\begin{equation}\label{piu}
    \text{Pixel } IU=\frac{TP}{TP+FP+FN}
\end{equation}
where TP is the global sum of line TPs, FP is the global sum of line FPs, and FN is the global sum of line FNs.

Line IU is measured at line level. For each matching pair, line precision and line recall are:
\begin{equation}\label{lineprecision}
    \text{Line precision}=\frac{\text{line } TP}{\text{line } TP + \text{line } FP}
\end{equation}
\begin{equation}\label{linerecall}
    \text{Line recall}=\frac{\text{line } TP}{\text{line } TP + \text{line } FN}
\end{equation}

Accordingly, line IU is:
\begin{equation}\label{liu}
    \text{Line } IU=\frac{\text{CL}}{\text{CL+ML+EL}}
\end{equation}
where CL is the number of correct lines, ML is the number of missed lines, and EL is the number of extra lines. 

For each matching pair:
\begin{itemize}
    \item A line is correct if both, the line precision and the line recall are above the threshold value.
    \item A line is missed if the line recall is below the threshold value.
    \item A line is extra if the line precision is below the threshold value.
\end{itemize}
\begin{figure}[h]
\centering
\includegraphics[width=7cm]{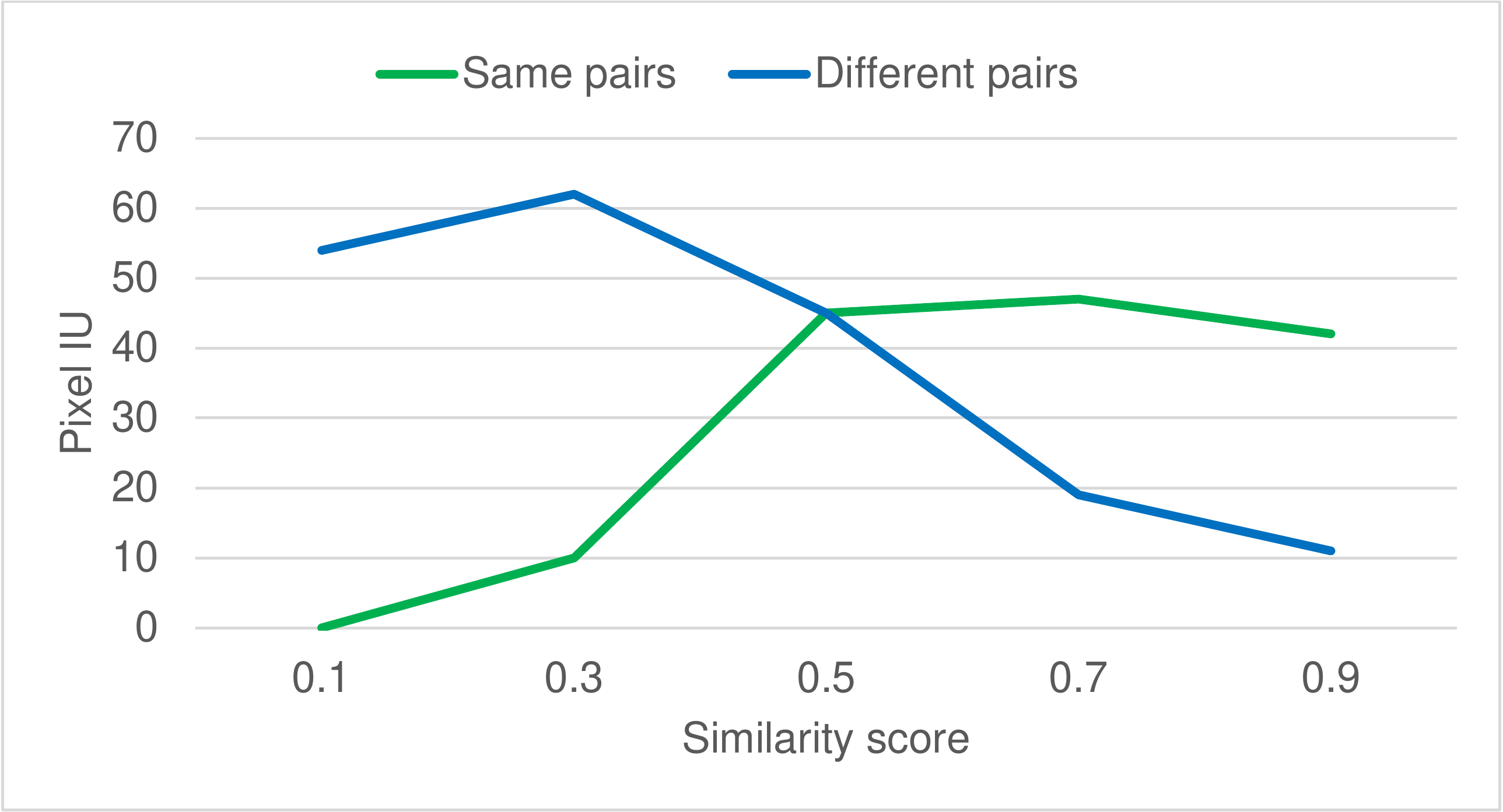}
\caption{Effect of different similarity score values for similar pairs and different pairs.}
\label{ablation}		
\end{figure}
\subsection{Effect of similarity scores}
\label{effect_of_similarity_scores}
As we describe in \sectionautorefname~\ref{datapreparation} two patches are most similar when similarity score $s \to 1$ and most different when $s \to 0$. We study this argument on a single page from VML-AHTE dataset. First, we fix $s \ge 0.5$ for similar pairs and report the effect of $t$ in $s \le t$ for different pairs. Then, we fix $s \le 0.5$ for different pairs and report the effect of $t$ in $s \ge t$ for similar pairs. The effect of different similarity score values can be observed in \figurename~\ref{ablation}.

\subsection{Effect of patch size}
\label{effect_of_patch_size}
\begin{figure}[h]
\centering
\includegraphics[width=6cm]{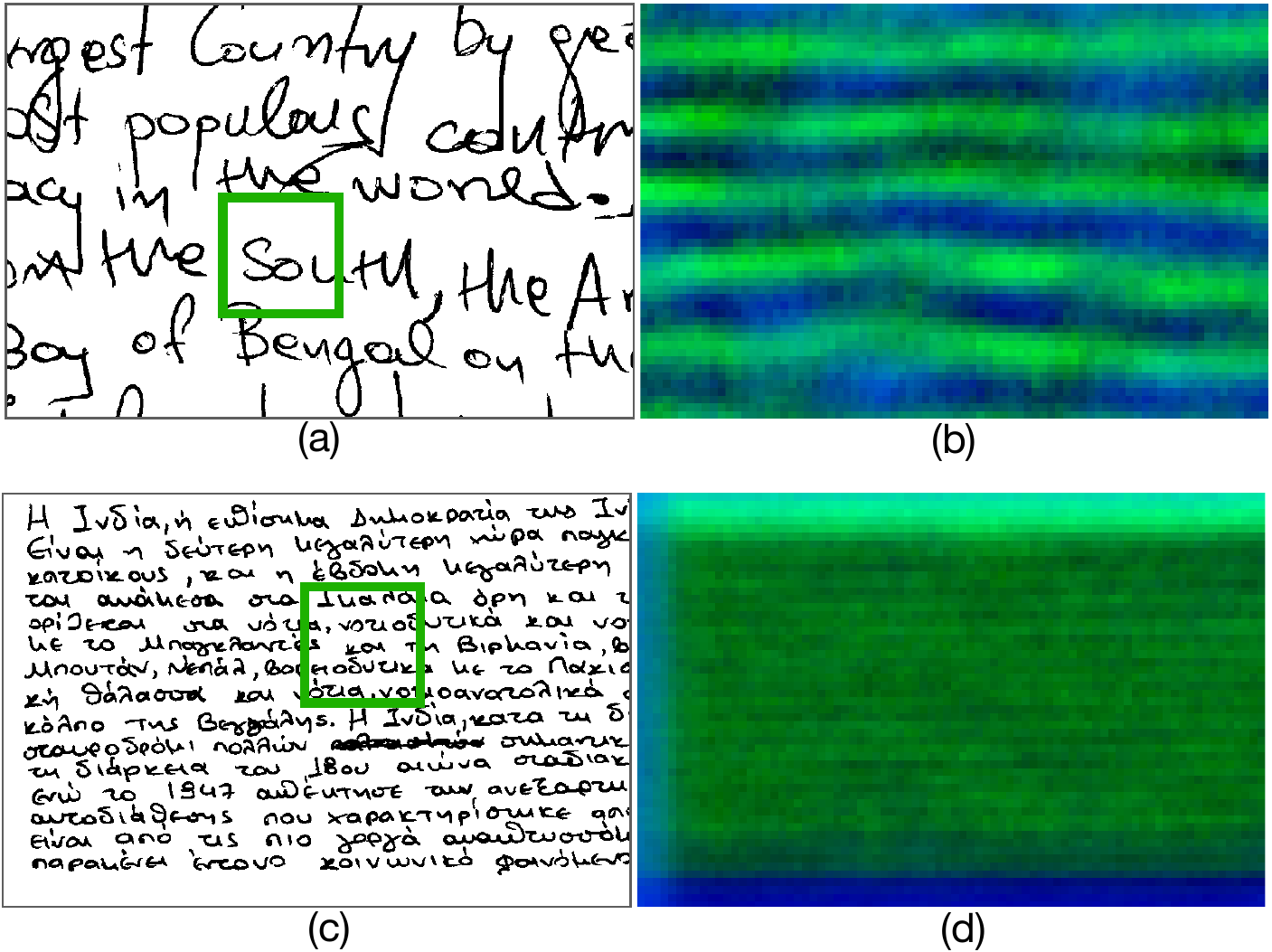}
\caption{Two sample document images from ICFHR 2010 dataset with very different text line heights (a) and (c). Same patch size can detect the text lines when its height is approximately 3 times the text line height (b) and can not detect the text lines when it spans several text lines together (d).}
\label{patch_size}
\end{figure}

\begin{figure}[h]
\centering
\includegraphics[width=5cm]{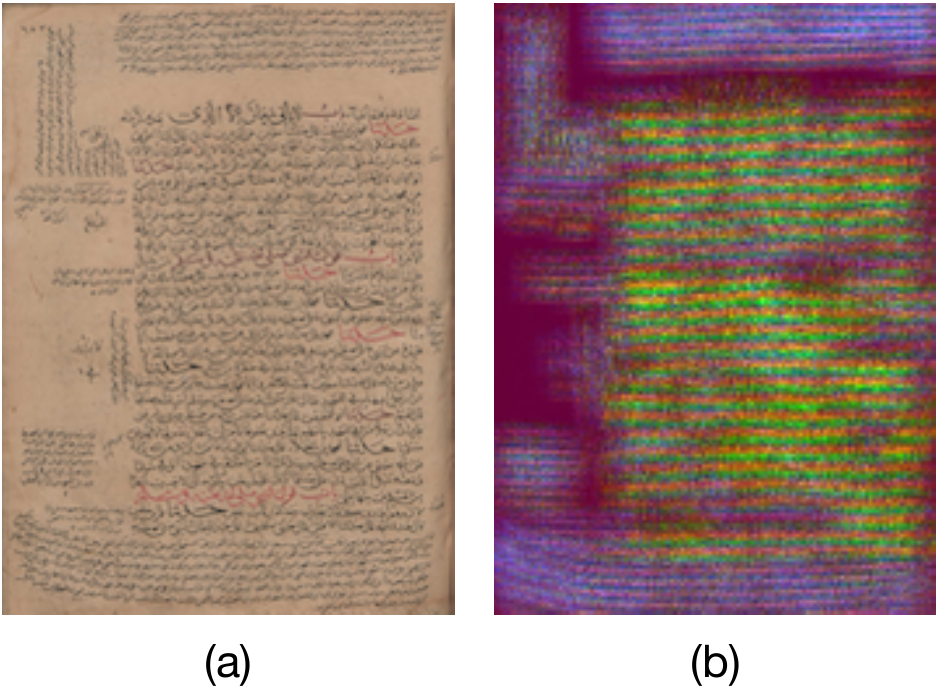}
\caption{A sample document image with heterogeneous text line heights (a). The pseudo-RGB output from the proposed method (b).}
\label{heterolines}
\end{figure}

We have found that the patch size is a critical value for the effective performance of the algorithm. If the documents in a dataset contain text lines that have severely different heights, then the algorithm does not produce good results, as shown in \figurename~\ref{patch_size}. This inaccuracy is caused by the inappropriate height estimation, which is taken as three times the average text line height in the documents. On the other hand we observe that a constant patch size can detect text lines with slightly different heights (\figurename~\ref{heterolines}).

\section{Results}
\label{results}
\subsection{Results on VML-AHTE dataset}
We compare our results with those of supervised methods, Mask-RCNN and FCN+EM and Human+EM. Mask-RCNN method is fully supervised using the pixel labels of the text lines. The advantage of this method is that it directly outputs pixel labels of text lines and does not need an additional procedure. FCN+EM method is also fully supervised but using blob lines that pass over the text lines. It uses EM framework to extract the pixel labels of text lines. Human+EM method is supervised by blob lines that are drawn by a human and uses EM framework to extract the pixel labels of text lines.

The comparison in terms of ICDAR 2013 metrics are reported in \tableautorefname~\ref{ahte_icdar2013metric_results}.
\begin{table}[h]
\centering
\caption{DR, RA and FM values on VML-AHTE dataset.}
\label{ahte_icdar2013metric_results}
\begin{tabular}{@{}rrrr@{}}
\toprule
Method & DR & RA & FM \\ \midrule
\textbf{Unsupervised}\\
UTLS& 93.62   &  \textbf{93.95}   & 93.78   \\
\hline
\textbf{Supervised}\\
Mask-RCNN  & 84.43   &  58.89   & 68.77   \\  
FCN+EM     & \textbf{95.55}   &  92.80   & \textbf{94.30}   \\
Human+EM   & 95.15   &  95.15   & 95.15   \\ \bottomrule
\end{tabular}
\end{table}

The comparison in terms of ICDAR 2017 metrics are reported in \tableautorefname~\ref{ahte_icdar2017metric_results}.
\begin{table}[h]
\centering
\caption{Line IU and pixel IU values on VML-AHTE dataset.}
\label{ahte_icdar2017metric_results}
\begin{tabular}{@{}rrr@{}}
\toprule
Method & Line IU & Pixel IU \\ \midrule
\textbf{Unsupervised}\\
UTLS& \textbf{98.55}   &  88.95   \\
\hline
\textbf{Supervised}\\
Mask-RCNN  & 93.08   &  86.97   \\ 
FCN+EM     & 94.52   &  \textbf{90.01}   \\
Human+EM   & 99.29   &  91.49   \\ \bottomrule
\end{tabular}
\end{table}

On VML-AHTE dataset, UTLS successfully learns and discriminates between the text lines and the spaces among text lines. Moreover it outperforms all the supervised methods in terms of RA and line IU, and is competitive in terms of the other metrics. The error cases arise from few number of touching blob lines. Such errors can easily be eliminated but this is out of the focus of this paper.

\subsection{Results on ICDAR 2017 dataset}
The second evaluation is carried out on the Task 3 of ICDAR 2017 Competition on Layout Analysis for Challenging Medieval Manuscripts. Within the Task 3 only the main body lines are in the scope of interest. We run our algorithm on pre-segmented text block areas by the given ground truth. Hence, we can compare our results with unsupervised System 8 and System 9 which are based on layout analysis prior to text line segmentation. 
The comparison in terms of ICDAR 2017 metrics are reported in \tableautorefname~\ref{icdar_icdar2017metric_results}. 

Main challenge in this dataset for UTLS is the wide spaces between the words in a text line. The wider the space between words the much likely the algorithm detects it as a space instead of a text line, which in turn leads to over segmentation.
\newcolumntype{G}{>{\centering\arraybackslash}m{0.22\textwidth}}
\newcolumntype{P}{>{\centering\arraybackslash}m{0.105\textwidth}}
\renewcommand\arraystretch{1.2}
\begin{table*}[h!]
    {\centering
    \caption{Results for the ICDAR 2017 competition on layout analysis for challenging medieval manuscripts. Line IU and Pixel IU results for Task 3}
    \begin{tabular}{l P P | P P | P P}
        \toprule
        \multicolumn{1}{l}{} &\multicolumn{2}{G|}{CB55} &  \multicolumn{2}{G|}{CSG18} & \multicolumn{2}{G}{CSG863}\\
        \hline
        \multicolumn{1}{c}{} & LIU & PIU & LIU & PIU & LIU & PIU\\
        \hline
        \textbf{Unsupervised}   &   &   &   &   &   &  \\
        UTLS                    & 80.35 & 77.30 & 94.30 & 95.50 & 90.58 & 89.40\\
        System-8 (CIT-lab)      & \textbf{99.33} & 93.75 & 94.90 & 94.47 & 96.75 & 90.81\\
        System-9+4.1 (DIVA+MG1) & 98.04 & \textbf{96.67} & \textbf{96.91} & \textbf{96.93} & \textbf{98.62} & \textbf{97.54}\\
         \toprule
    \end{tabular}
    \label{icdar_icdar2017metric_results}
    }
    LIU denotes Line IU and PIU denotes Pixel IU
\end{table*}

\subsection{Results on ICFHR 2010 dataset}
This dataset contains very heterogeneous text lines with excessively different heights, interline spaces, and skews. The comparison on ICFHR 2010 dataset using ICDAR 2013 metrics are reported in \tableautorefname~\ref{icfhr2010_icdar2013metric_results}.

Main challenge in this dataset for UTLS is the severely different text line heights. The algorithm can not detect the text lines with heights that are very greater than or very less than the patch height (\figureautorefname~\ref{patch_size}).
\begin{table}[h]
\centering
\caption{DR, RA and FM values on ICFHR 2010 dataset.}
\label{icfhr2010_icdar2013metric_results}
\begin{tabular}{@{}rrrr@{}}
\toprule
Method & DR & RA & FM \\ \midrule
\textbf{Unsupervised}\\
UTLS   & 73.22 &  72.38 & 72.36   \\
Winner & \textbf{97.54} &  \textbf{97.72} & \textbf{97.63}   \\  
\hline
\textbf{Supervised}\\
\cite{diem2013text}  & 97.18  &  96.94 & 97.06   \\ \bottomrule
\end{tabular}
\end{table}

\section{Conclusion}

We have presented an unsupervised text line segmentation method UTLS, trained to discriminate the text lines from the spaces between text lines. UTLS learn feature representations that are comparable or superior to other models trained with full supervision. The method is convenient in terms of average prediction time per page using a single Intel Xeon GPU (\tableautorefname~\ref{runtimes}).

The algorithm is very effective in detecting cramped and crowded text lines with nearly constant heights, interline spaces and interword spaces. However heterogeneity of aforementioned features decreases the performance of UTLS significantly.
\begin{table}[h]
\centering
\caption{Average prediction run times per page for each dataset in terms of minutes.}
\label{runtimes}
\begin{tabular}{@{}rrrr@{}}
\toprule
                              & VML-AHTE & ICDAR 2017 & ICFHR 2010 \\ \midrule
Average run time per page     & 2.62     &  2.20      & 1.49   \\  \bottomrule
\end{tabular}
\end{table}

\section*{Acknowledgment}
The authors would like to thank Gunes Cevik for data preparation. This work was supported by Frankel Center for Computer Science at Ben-Gurion University.

\bibliographystyle{IEEEtran}
\bibliography{IEEEabrv,xref.bib}
\end{document}